%% file: main.tex
\documentclass[11pt]{article}
\usepackage[a4paper, total={6in, 8in}]{geometry}
\usepackage{hyperref}
\hypersetup{colorlinks,
            linkcolor=blue,
            citecolor=blue,
            urlcolor=magenta,
            linktocpage,
            plainpages=false}

\usepackage{longtable}

\usepackage{graphicx}
\usepackage{booktabs}
\usepackage{subfig}
\usepackage{wrapfig}

\usepackage[load-configurations=version-1]{siunitx} 
\usepackage{natbib}

\usepackage{amsmath}
\usepackage{amssymb}
\usepackage{amsthm}
\usepackage{mathtools}
\usepackage{dsfont}
\usepackage[capitalise]{cleveref}
\usepackage{tikz}
\usepackage{multirow,multicol}
\usepackage{bm}
\usepackage{forest}
\usepackage{tikz}
\usepackage{array}
\usepackage[most]{tcolorbox}
\usepackage[page,header]{appendix}
\usepackage{colortbl}

\usepackage{algorithm}
\usepackage{algorithmic}

\usetikzlibrary{trees,bayesnet,arrows,automata,positioning}

\usepackage{thmtools}

\tcolorboxenvironment{lemma}{enhanced jigsaw,colback=gray!20,drop shadow}

\declaretheorem{proposition}
\tcolorboxenvironment{proposition}{enhanced jigsaw,colback=gray!20,drop shadow}

\tcolorboxenvironment{remark}{enhanced jigsaw,colback=gray!20,drop shadow}

\declaretheorem{definition}
\tcolorboxenvironment{definition}{enhanced jigsaw,colback=white,drop shadow}

\DeclarePairedDelimiter{\ceil}{\lceil}{\rceil}
\definecolor{Gray}{gray}{0.85}
\newcolumntype{a}{>{\columncolor{Gray}}c}

\usepackage[affil-sl,auth-sc]{authblk}

\definecolor{Gray}{gray}{0.85}
\newcolumntype{a}{>{\columncolor{Gray}}c}

\DeclareMathOperator*{\argmax}{argmax}

\DeclareMathOperator{\HQS}{HQS}
\newcommand{\stay}{a_s}
\newcommand{\go}{a_d}
\DeclareMathOperator{\similarity}{sim}
\newcommand{\guid}{\eta}

\newcommand{\tree}{\mathcal{H}}
\newcommand{\state}[2]{\left\langle #1, #2 \right\rangle}

\title{Partially Observable Markov Decision Process Modelling for Assessing Hierarchies}

\author[1,2]{Weipeng Huang\thanks{weipeng.huang@insight-centre.org}}
\author[1,3]{Guangyuan Piao\thanks{guangyuan.piao@insight-centre.org}}
\author[1,2]{Raul Moreno\thanks{raul.moreno.salinas@gmail.com}}
\author[1,2]{Neil Hurley\thanks{neil.hurley@insight-centre.org}}
\affil[1]{Insight Centre for Data Analytics}
\affil[2]{School of Computer Science, University College Dublin, Dublin, Ireland}
\affil[3]{Data Science Institute, National University of Ireland, Galway, Ireland}

\begin{document}

\maketitle

\input{abstract}

\input{intro}
%
\input{relatedwork}

\input{modeldiscussion}
\input{policy}
\input{experiment}
\input{conclusion}


\bibliographystyle{abbrvnat}
\bibliography{evalmodel}

\input{appendix}

\end{document}

%% file: abstract.tex
\begin{abstract}
Hierarchical clustering has been shown to be valuable in many scenarios.
Despite its usefulness to many situations, there is no agreed methodology on how to properly evaluate the hierarchies produced from different techniques, particularly in the case where ground-truth labels are unavailable.
This motivates us to propose a framework for assessing the quality of hierarchical clustering allocations which covers the case of no ground-truth information.
This measurement is useful, e.g., to assess the hierarchical structures used by online retailer websites to display their product catalogues.
Our framework is one of the few attempts for the hierarchy evaluation from a decision theoretic perspective.
We model the process as a bot searching stochastically for items in the hierarchy and establish a measure representing the degree to which the hierarchy supports this search.
We employ Partially Observable Markov Decision Processes (POMDP) to model the uncertainty, the decision making, and the cognitive return for searchers in such a scenario.
\end{abstract}

%% file: intro.tex
\section{Introduction}
\label{sec:intro}
Hierarchical clustering (HC) analysis has been applied in many E-commerce and scientific applications.
The process generates a collection of nested clusters that group the data in a connected structure~\citep{balcan2014}, forming a hierarchy.
The hierarchy is represented by a tree data structure where each node contains a number of data items. Such a structured organisation of the items is useful for tasks such as efficient search. Searchers can navigate through the catalogue of items by choosing a path through the hierarchy and can thus avoid the cost of a linear search through the entire catalogue of items. If the hierarchy is well organised into coherent clusters, then finding the correct path to the required item is easy. Clearly, there are  multiple ways of clustering the same items and organising their hierarchies. However, evaluating the quality of a hierarchy still needs more substantial study, especially for data that lacks a ground-truth hierarchy. The evaluation of hierarchical structures is the focus of this paper.

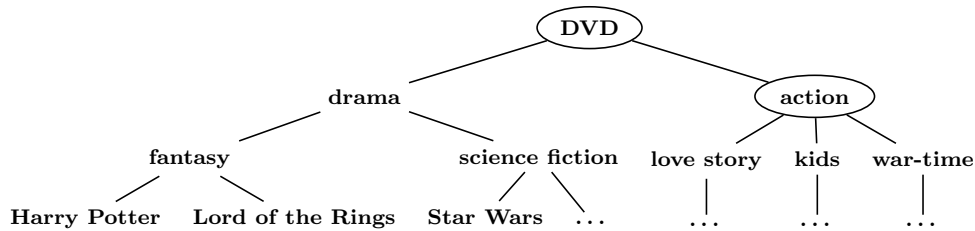
\begin{figure}
\centering
\scriptsize
\begin{forest}
  for tree={
    fit=band,
    line width=.6pt,edge={line width=.6pt},
  }
  [\textbf{DVD}, draw, ellipse
    [\textbf{drama},
      [\textbf{fantasy},
        [\textbf{Harry Potter}]
        [\textbf{Lord of the Rings}]
      ]
      [\textbf{science fiction}
        [\textbf{Star Wars}]
        [\textbf{\dots}]
      ]
    ]
    [\textbf{action},draw, ellipse
      [\textbf{love story} [\textbf{\dots}]]
      [\textbf{kids} [\textbf{\dots}]]
      [\textbf{war-time} [\textbf{\dots}]]
    ]
  ]
\end{forest}
\caption{An example which illustrates a hierarchy that confuses the searchers given that the target is a Harry Porter DVD}
\label{fg:close-example}
\end{figure}

We consider on-line retailers and the customers who access them, as an example.
When looking for specific items, customers are in essence navigating the hierarchies hosted by the website.
Different clusters and hierarchies will probably provide divergent levels of user experience with respect to searching and navigating, even for the same users.
For example, imagine that a female jeans is mistakenly placed in a branch of a certain hierarchy, labelled ``clothes $\rightarrow$ trousers $\rightarrow$ jeans $\rightarrow$ males''.
The user will expect that the jeans are contained along the branch ``clothes $\rightarrow$ trousers $\rightarrow$ jeans $\rightarrow$ females'', and thus will surely fail to retrieve the required item in that hierarchy.
As another example,~\cref{fg:close-example} shows that a searcher may be confused and possibly search for a Harry Potter DVD in the wrong route.
Is it more so an action movie than a drama movie?
In this second example, it may well be the case that each cluster contains a coherent set of items, but their hierarchical organisation makes it difficult for a searcher to choose the correct path.  Such observations motivate an evaluation function that accounts for the structure as much as the item to cluster assignments.

In the scenario that we consider here, all items stored in the sub-tree rooted at a node in the hierarchy are available for consideration by a searcher who stops at that node. For instance, a searcher stopping at the \emph{drama} node could consider in turn all the DVDs, ``Harry Potter'', ``Lord of the Rings'', ``Star Wars'', and so on. The further the searcher descends in the hierarchy, the fewer items that need to be considered once the searcher stops to search, thus increasing search efficiency, provided that a correct path in the hierarchy is taken.

There are many evaluation functions already proposed to measure the quality of regular (non-hierarchical) clustering results, with or without ground-truth data. Any of these measures can be applied to a HC, once an appropriate cut of the hierarchy is chosen. Such measures do give feedback on the coherence of the clusters and/or their agreement with a ground truth.
However, there is relatively little research that has proposed measures scoring the structural organisation of the hierarchy.
This is critical in order to properly evaluate hierarchies in the context that we have in mind.
In particular, one can extend it to perform the hyper-parameter tuning for a number of HC algorithms.
As demonstrated in~\citet{kobren2017hierarchical}, many HC algorithms depend on hyper-parameters and might possibly be further improved with our function.

\paragraph{Problem Statement}
Let us consider a set of items $X=\{x_n\}^N_{n=1}$.
HC is a set of nested clusters of the dataset arranged in a tree $\tree$.
Each node of the tree has associated with it a sub-set or cluster of the collection, and the children of any node are associated with a partition of the parent's cluster.
The root node is associated with the entire collection, and if an item belongs to any particular node in $\tree$, it also belongs to all of its ancestor nodes.
The goal is to seek a function mapping $\tree$ to a real value, which reflects the quality of the hierarchical arrangement, from the point of view of supporting efficient search for a \emph{target} item in the collection.
In summary, this work aims to arrive at a quantitative quality measure for a given hierarchical arrangement of a catalogue of items, where in a typical use-case, the items are products in a large catalogue offered by an on-line retailer.
Our scenario is that of a search bot seeking a specific target item and so we develop the measure by modelling a simplified search process but one which is sufficient to capture the important features of the hierarchy that determine its suitability to support efficient search.

It is intuitive to model this process of decision making under
uncertainty as a Markov Decision Process (MDP).
MDPs are concerned with the problem of determining a decision policy that optimises the cumulative reward obtained when applying this policy over some time horizon.
\citet{Moreno2017} proposed such an MDP to tackle the problem of evaluating hierarchies. In that model, the quality of the hierarchy corresponds to the expected cumulative reward that a search bot will obtain when searching for target items, using a particular search policy, where the reward is a function of whether and how efficiently the target is found.
Notice that, this quality measure depends not only on the hierarchy itself, but also on the policy used to determine the action at each decision point.

\paragraph{Contribution}
We propose a novel framework extending the MDP model with a reasonable search policy proposed, which provides a solid model for measuring the efficiency of item retrieval in the hierarchy.
Our framework explicitly models the searcher's lack of knowledge about the environment.
This leads us to extend from an MDP model, to a Partially Observable MDP (POMDP) model.
We coin the measure Hierarchy Quality for Search (HQS).


%% file: relatedwork.tex
\section{Related Works}
\label{sec:related}

When assessing general clustering results, one can appeal to measuring the results with or without the ground-truths~\citep{Liu13, Steinbach00}.
A ground-truth hierarchy should contain multiple layers assigning nodes to each level of the hierarchy.
Thus, we do not consider the Dendrogram Purity (e.g., used in \citet{heller2005bhc,kobren2017hierarchical}) as it evaluates a hierarchy with regard to only one layer of cluster assignment.
There are many tools for evaluating the quality of the clustering but they lack the capability to evaluate the hierarchical arrangement~\citep{Johnson2013}.

\cite{cigarran2005evaluating} proposed a prototype considering the content of the cluster, the hierarchical arrangement and the navigation cost.
Unfortunately, it is hard to develop this idea as neither detailed procedures nor experiments were presented in this short paper.
More recently,~\cite{Johnson2013} proposed the Hierarchical Agreement Index (HAI), which borrows its concept from the Rand Index, to compare the structure to a ground-truth hierarchy.
\cite{Moreno2017} introduced the concept of using MDPs to model the evaluation.
They observed that HC measures have hardly addressed the need to understand the convenience for the search and navigation efficiency provided by a hierarchy, accounting for the cognitive cost of choosing a correct path at each branch of the hierarchy.
However, it is a prototype rather than a finished product.
Our work extends these ideas to a POMDP~\citep{Kaelbling1998} model, which improves the previous work by 1) modelling the uncertainty in the search process with belief states; 2) specifying a policy to solve the problem so that the strong assumptions (the bot knows the position of the items universally) in the previous work can be relaxed.
We consider this work an important contribution to the topic of evaluating hierarchical structures with no help from ground-truths.

There exist abundant research addressing the issue of acting optimally in POMDPs, such as~\citep{cassandra1994,Boutilier1996,meuleau1999solve,Ross2008}, to name but a few.
Seeking a better policy is not the main focus in the paper; hence, we adopt the existing (on-line) policies.
On the other hand, we find the works~\citep{shani05,zheng2018drn,Eugene2019} employ Reinforcement Learning (RL) to model the user behaviours for improving the performance in recommender systems, which are inspiring.
These are somewhat close to our work while in a different application domain.

%% file: modeldiscussion.tex
\section{POMDP Specification}
\label{sec:pomdp}
We develop a POMPD model of a bot searching a hierarchy, where, at each node, it must make the decision to search or to descend further. The bot cannot backtrack, once a descent step has been performed.
Simply put, a POMDP is an MDP with additional observations, belief state and belief estimator~\citep{cassandra1994,Kaelbling1998}.
Fixing the search target of the bot as item $x$, we model the search of the hierarchy as the POMDP  $\mathcal{P}_x = \langle {S}, {A}, {T}, {R}, Z, O, \gamma \rangle$, such that
\begin{itemize}
\item
$S$ is the finite state space.
\item
${A}$ is the set of possible actions.
\item
$T: S \times A \times S \mapsto [0, 1]$ is the transition function where $T(s, a, s') \triangleq p(s' \mid s, a)$, represents the probability of moving to state $s'$ from state $s$ using action $a$.
\item
${R}: S \times A \times S \mapsto \mathbb{R}$ assigns the expected immediate reward, $R(s,a, s')$, to the resulting state $s'$ after the bot selects action $a$ in state $s$.
\item
$O$ is the observation set.
\item
$Z: S \times A \times O \mapsto [0,1]$ is the probability function for the observations.
We have $Z(s, a, o) = p(o \mid a, s)$ which reads as the probability of observing $o \in O$ when reaching the state $s$ through action $a$.
\item
$\gamma$ is the discount factor. In our model, it is sufficient to set $\gamma=1$ and hence we will omit further explanation about this part.
\end{itemize}

\subsubsection*{States}
POMDPs use states to represent the status of the search at a particular point during the search process.
Our model defines that each state is a joint event consisting of 1) the physical location (node) $c$ of the bot, and 2) a \emph{boolean} variable indicating if the bot is in the right path:
\begin{align}
  S = \{\state{\emptyset}{0},\state{\emptyset}{1}\} \cup  \bigcup_{c\in\tree}\{\state{c}{0},\state{c}{1} \} \, ,
\end{align}
where $\emptyset$ is the terminal point reached after the bot chooses to search.
Accordingly, $\state{\emptyset}{1}$ is the state representing that the bot stops and the stopping node contains the item, likewise  $\state{\emptyset}{0}$ depicts the opposite case.

\subsubsection*{Actions}
We first discuss the  \emph{guidance function} concept, which is a core component in our model.
Then, we specify the action set.

\paragraph{Guidance Function}
A guidance function $\guid: 2^X \times 2^X \times X \to [0,1]$  provides the search bot with evidence as to the correct path on which to search for an item $x \in X$.
Specifically, $\guid(c,c',x)$ represents the bot's estimate that the target $x$ is contained in a cluster $c'$, given that it is in cluster $c$, for $(c,c') \in \tree$.
We have
\begin{align*}
  \forall c' \notin \mathcal{C}(c): \guid(c, c',x) = 0 \qquad \mbox{and} \qquad \sum_{c' \in \mathcal{C}(c)} \guid(c, c',x) = 1 \,
\end{align*}
where $\mathcal{C}(c)$ returns the children of $c$.
The $\guid$ function summarises the information (prior domain knowledge) available in the hierarchy which will guide the bot's behaviour.
It is represented as a discrete probability density function.
In particular, we write for any parent-child pair $(c, c') \in \tree$ that,
\begin{equation}
  \label{eq:eta}
  \guid(c, c',x) = p(x \in c' \mid x \in c) \triangleq
  \frac {\exp\{\similarity(x, c')/\delta\}} {\sum_{c'' \in \mathcal{C}(c)} \exp\{\similarity(x, c'')/\delta\}} \,
\end{equation}
where $\similarity(x, c)$ is the similarity function between the item $x$ and the cluster $c$.\footnote{In the rest of the paper, we omit $x$ in the guidance function, since the discussions about the guidance function will always concentrate on one specific item.}
Additionally, $p(x \in c' \mid x \notin c)$ is always $0$ as it is impossible to retrieve the item in a node whose parent does not contain it.
Leaving the choice of similarity function open adds flexibility to the measure for adjusting to diverse datasets and applications.
The $\guid$ function should return a higher probability score for the node $c'$ that is ``closest'' to $x$ among the siblings.
The parameter $\delta$ is a \emph{temperature} parameter in the Boltzmann function.
We disclose more insights about this function in the supplemental materials about how to control the function as the path goes deeper.

\paragraph{Action Set}
As our measure should represent the quality of the hierarchy for a typical searcher, we capture the possible search behaviours by modelling that the bot searches stochastically.
Nevertheless, the measure focuses on rational search behaviour, rather than on fully random search.
The proposed framework allows us to capture this rationality by exposing part of the search behaviour to rational decision making.
Hence, we impose randomness in the manner in which children are selected by a searcher and expose only the decision of whether or not to explore the structure of the hierarchy to the bot's decision-making logic.
We design the bot to move randomly according to the guidance function $\guid(\cdot)$.
Specifically, the bot moves to a child $c'$ with probability $\guid(c,c')$, i.e. based on the probability of the item being stored at the target location.
With the above rationale, the action set contains only two actions: descend $\go$ and search $\stay$, corresponding to the choices of descending the hierarchy to another level, or stopping and searching for the target at the current node.

\subsubsection*{Transitions}
The transition is fixed and the bot can fully observe the outcome, when the bot searches.
That is, the bot knows whether it is in $\state{\emptyset}{0}$ or $\state{\emptyset}{1}$ reaching the state $\emptyset$ when applying action $\stay$.

For the descent action, the bot estimates the transition i.e. it computes $\hat{T}$ such that $\hat{T}(s, a, s') = \hat{p}(s' \mid s, a)$.
In regular MDP and POMDP problems, the unknown transitions are handled by the RL methodology, such that the bot can explore by choosing certain actions and can approximate the transition probabilities by the ratio of the number of ending states over all the attempts~\citep{duff2002optimal,ng2012bayes}.
In our task, we have to design for bot's transition distribution rather than to learn them from the data.
Thus, the ``domain knowledge'' should play the role for driving the bot to succeed or fail in its task.

Let $g_c$ be the \emph{boolean} variable associated with a node $c$ encapsulated in a state $s$. We decompose the transition probability of applying $\go$ for the various cases.
For $c' \in \mathcal{C}(c)$, $s = \state{c}{g_c}$ and $s' = \state{c'}{g_{c'}}$,
\begin{align*}
\hat{p}(s' \mid s, \go)
&= p\left(\state{c'}{g_{c'}} \mid \state{c}{g_{c}}, \go \right) = p\left(c' \mid c, \go \right) \hat{p}\left(g_{c'} \mid g_{c}\right) = \guid(c,c') \hat{p}\left(g_{c'} \mid g_{c}\right)\,
\end{align*}
where the last step follows from the stochastic descent process described in the previous section.
The $\guid$ function is also used for the estimate $\hat{p}(g_{c'} \mid g_c)$ such that, $\hat{p}(g_{c'} =1 \mid g_c = 1) = \guid(c, c')$.
Overall, for $\go$, we get, for all $c \in \tree$, $c' \in \mathcal{C}(c)$:
\begin{align}
\label{eq:trans}
p(\state{c'}{1} \mid \state{c}{1}, \go) &= \guid(c, c') \hat{p}(g_{c'} =1 \mid g_c = 1) = \guid(c, c')^2 \nonumber \\
p(\state{c'}{0} \mid \state{c}{1}, \go) &= \guid(c, c') \hat{p}({g}_{c'} = 0 \mid g_c = 1) = \guid(c, c')(1 - \guid(c, c')) \nonumber \\
p(\state{c'}{1} \mid \state{c}{0}, \go) &= \guid(c, c') \hat{p}(g_{c'} = 1 \mid {g}_c = 0) = 0 \nonumber \\
p(\state{c'}{0} \mid \state{c}{0}, \go) &= \guid(c, c') \hat{p}(g_{c'} = 0 \mid {g}_c = 0) = \guid(c, c')  \, .
\end{align}

\subsubsection*{Rewards}
In the POMDP, only when the bot stops and searches is a non-zero reward obtained; navigating earns $0$ reward.
Moreover, searching in a node with fewer items gives the bot a higher cognitive reward which motivates the bot to traverse down the tree.
In particular, we have
\begin{align*}
  R(s, a, s') =
  \begin{cases}
    0       & a = \go \\
    r( c )       & s' = \state{\emptyset}{1} \\
    -1      & s' = \state{\emptyset}{0}
  \end{cases}
  \quad and \quad
  r(c) & = 1 - \left(e^{|c| / N} - 1 \right) / (e-1)
\end{align*}
where $c$ is the location encapsulated in $s$.
We assign $-1$ to searching at a wrong node.
One can customise $r(\cdot)$ as long as it is monotonically decreasing with the size of the input cluster to be searched.
The selected $r(c)$ approximates the ease of searching within a set of items where $|c|$ is the number of items at $c$.

\subsubsection*{Observations and Observation Probabilities}
The observations after the bot descends are its own location in the hierarchy, the children, and some abstract summaries from the children, etc.
This information is fixed and unique for each location that the bot arrives at.
Hence one can integrate it into one indexed observation that directly affects our belief update function.
Recall that $Z(s, a, o) = p(o \mid s, a)$.
It is trivial to see $p(o \mid \state{c'}{g_{c'}}, \go) = \mathds{1}\{o = c'\}$, i.e. the bot can detect its exact location.
After performing $\stay$, the observations are whether or not the target is found in the current node.

\subsubsection*{Belief Update}
As states are not fully observable, the bot maintains its belief state $b$ as a probability function at any specific time.
Belief updates can be written using the \emph{belief update} function, $\tau$, where $b' = \tau(b,a,o)$, is the updated belief when observation $o$ is made after action $a$ is applied in belief state $b$.
By Bayes' Theorem,
\begin{align}
\label{eq:bf-update}
  b'(s') &= p(s' \mid a, o, b) = \frac {p(o \mid s', a) \sum_s p(s' \mid s, a) b(s)} {p(o \mid a, b)} \, .
\end{align}

Considering that $\go$ is performed at $c$, then for each possible child $c' \in \mathcal{C}(c)$, such that the new state is $s'=\state{c'}{0}$ or $s'=\state{c'}{1}$, we have
\[
  b'\left(s'\right) \propto \mathds{1}\{o = c'\} \sum_s p(s' \mid s, \go) b(s) \, .
\]
To be more specific,
\begin{align}
b'\left(\state{c'}{0}\right)
&\propto \mathds{1}\{o = c'\} \sum_s p(\state{c'}{0} \mid s, \go) b(s) \nonumber \\
&= p(\state{c'}{0} \mid \state{c}{1})b(\state{c}{1}) + p(\state{c'}{0} \mid \state{c}{0})b(\state{c}{0}) \nonumber \\
&=\guid(c, c') [1 -\guid(c, c')b(\state{c}{1})] \,.\label{eq:b0} \\
b'\left(\state{c'}{1}\right)
&\propto
\guid(c, c')^2 b\left(\state{c}{1}\right) \label{eq:b1} \, ,
\end{align}
where~\cref{eq:b1} is obtained in a similar way to~\cref{eq:b0}.
As discussed earlier, for other states, $s''$, not related to $c'$, $b'(s'')=0$.
To compute the normalising constant $p(o \mid b, \go)$, when $a = \go$, we have that
\begin{align*}
p(o=c' \mid b, \go) = \guid(c, c') \left[1 - \guid(c,c')b(\state{c}{1}) + \guid(c, c') b(\state{c}{1})\right]
&= \guid(c, c') \, ,
\end{align*}
which gives a final belief update rule of
\begin{align}
\label{eq:beliefupdate}
b'\left(\state{c'}{1}\right)
&= \guid(c, c') b\left(\state{c}{1}\right) \qquad  \mbox{and} \qquad b'\left(\state{c'}{0}\right) = 1 - \guid(c, c') b\left(\state{c}{1}\right) \, .
\end{align}
Note that $p(o \mid b, \stay) = 1$ for reaching the terminal state, such that $o = \emptyset$.

Let us denote by $s = s_{t}$ the state reached after $t$ update steps, with similar sub-scripting for $b$ and $c$.
The root node is $c_0$. Throughout our analysis, we assume that the target $x$ is certainly contained inside the hierarchy and hence, the bot always starts in state $\state{c_0}{1}$.
It follows that $ b(\state{c_0}{1})=1$.

By induction, we arrive at a simple expression for the belief state when the search has reached node  $c_T$ at time $T$ where $T > 0$, $b_T(\state{c_{T}}{1}) = \prod_{t = 1}^{T} \guid(c_{t-1}, c_t)$ and $ b_T(\state{c_{T}}{0}) = 1 - \prod_{t=1}^{T} \guid(c_{t-1}, c_t)$ while $b_T(s)=0$ for all other states $s$.
That is, once the bot reaches a node $c$, the belief state of $\state{c}{1}$ is simply the probability  of reaching  this node from the root, and that of $\state{c}{0}$ is the residual, $1-b_T(\state{c}{1})$.
Furthermore, since there is no backtracking, there is exactly one belief state associated with each node $c \in \tree$, which, if $\{c_0,\dots,c_T=c\}$ is the unique path, is fully determined by the value $b_c \triangleq b_T(\state{c_{T}}{1})$ and we can simply write, for  $c' \in \mathcal{C}(c)$, the belief update as $b_{c'} =\guid(c,c') b_c$.


\subsubsection*{Value Function}
The most commonly used objective that drives policy selection in an MDP is the value $V$ (discounted long term reward), s.t. $V \triangleq \sum_{t=0} \gamma^t R_t$ where $R_t$ is the reward obtained at time stamp $t$.
Let use denote a policy by $\pi$.
Solving a POMDP aims to maximise the value function, $V^\pi(b)$, corresponding to the cumulative expected reward \emph{over the beliefs}~\citep{singh1994learning}.
In particular, $R_B(b, a) = \sum_{s} b(s)\sum_{s'}p(s'|s,a) R(s, a,s')$.
It follows that the belief value function of a policy $\pi$ is given by
\begin{align*}
V^\pi(b) = \sum_{a} \pi(b,a)\left[ R_B(b, a) + \gamma \sum_{s} b(s) \sum_{s'} p(s' \mid s, a) \sum_o p(o\mid s',a) {V^\pi}(\tau(b,a,o))\right]\,
\end{align*}
where $\pi(b,a)$ is the probability that action $a$ is selected in belief state $b$.
The optimal value $V^*(b)$ with a corresponding deterministic policy to choose action $a^*$ can be written as a Bellman equation:
\begin{align*}
{V}^*(b) &= \max_{a \in A} Q(b,a) \qquad a^* = \argmax_{a \in A} Q(b,a) \\
\text{where} \quad \quad Q(b,a)
&= R_B(b, a) + \gamma \sum_{o} p(o \mid b, a) {V^*}(\tau(b,a,o)) \,.
\end{align*}
This shows that the POMDP may be interpreted as an MDP over belief states with $p(o \mid b, a)$ the  transition probability for moving from belief state $b$ to belief state $\tau(b,a,o)$.

Let $T$ be the step at which a terminal state $\state{\emptyset}{0}$ or $\state{\emptyset}{1}$ is reached.
The cumulative reward is given by
\begin{align*}
  \sum_{t=0}^{T-1} R_t =
  \begin{cases}
    -1 & s_{T} = \state{\emptyset}{0} \\
    r(c_{T-1}) & s_{T} = \state{\emptyset}{1}
  \end{cases}\,.
\end{align*}
It is natural to examine the value $V^\pi$ achieved by the policy $\pi$ learned over the POMDP in the underlying MDP~\citep{singh1994learning}, that arises from the POMDP when the MDP states are fully observable.
$V^\pi$ may be interpreted as the value that an \emph{oracle} that knows the target's location would assign to the bot's policy.
We base the HQS measure on this oracle value, noting that the bot can only attempt to maximise the belief value $V^{\pi}(b)$.

\subsection{Example of the POMDP}
Due to the space limit, we provide a walk-through example in the supplemental document (Appendix A.) which shows how a POMDP will be constructed given a simple hierarchy.

\subsection{Hierarchy Quality for Search}
We measure the quality of the hierarchy as the oracle value, i.e. the value of this policy in the underlying MDP.
Since the reward function always outputs $-1$ whenever the policy leads to a wrong path, to compute this oracle value, we only need to focus on the correct path.
In particular, let $T$ denote the depth at which the bot chooses $\stay$ and let $\{c_0,c_1,\dots,c_T\}$ now denote the \emph{correct} path containing the target item $x$ to depth $T$. Then
\begin{align*}
V_x
&= r( c_T )\prod_{t = 1}^T \guid\left(c_{t-1}, c_t\right) + (-1) \times \left(1 - \prod_{t = 1}^T\guid(c_{t-1}, c_t) \right) = (r(c_T) + 1) \left(\prod_{t = 1}^T\guid(c_{t-1}, c_t) \right) - 1 \, .
\end{align*}

Intuitively, the bot moves randomly according to $\guid(\cdot)$, and when it stops at depth $T$, the probability of obtaining a positive reward is simply the probability that the guidance function led it along the right path.
Given $0 \le r(\cdot) \le 1$, we obtain $-1 \le V_x \le 1$ which matches our intuition that if there are fewer items stored in the node at which the bot stops and the uncertainty of reaching this node is smaller, the bot achieves a higher value.
The stopping point $T$, is the key output of the belief-based policy determining the oracle value.

Eventually, we define HQS as follows.
\begin{definition}[HQS]
The $\HQS$ is a function over the set of data $X=\{x_n\}_{n=1}^{N}$ and the hierarchy $\tree$, such that $\HQS(X, \tree) \triangleq \frac 1 N \sum_{n=1}^N V^{\pi}_{x_n}(\tree)$ where $V^{\pi}_{x_n}(\tree)$ is the long term expected reward for the bot to search item $x_n$ in $\tree$, obtained by taking policy $\pi$.
\end{definition}

%% file: policy.tex
\section{Solving the POMDP}
\label{sec:policy}
One can categorise POMDP solvers to off-line and on-line policies.
Off-line methods either learn the whole environment prior to determining the policy or learn it through RL.
On-line planners solve the POMDP in a real-time decision making context, focusing on the information in the current states~\citep{Ross2008}.
We appeal to on-line planners as they better simulate the situation of a bot making decisions as the search proceeds.
Also, off-line methods would make the process less tractable and thus unsuitable for the evaluation task.

One notable on-line Monte-Carlo based planner, POMCP~\citep{silver2010monte}, is able to earn a good underlying MDP value regardless of the quality of the belief states.
This \emph{contradicts} what we expect from the role of belief in our model.
If the guidance function is weak, then we expect that this should be reflected in a belief function that leads to a poor reward for the underlying MDP, as it should tend to lead the bot on a path not containing the target.
Hence, the bot is coded with the Real-Time Belief Space Search (RTBSS) planner to seek the policy~\citep{paquet2005online,paquet2005real,Ross2008}.
In RTBSS, at each decision point, the bot is restricted to learn the required information only of the immediate children, limiting it to only one layer down in the hierarchy.
This corresponds to \emph{two} look-aheads for an on-line planner, where the bot can compare stopping and searching at the current node, with that searching at the next layer after one descent.

\subsection{RTBSS}
We leave the details of the original RTBSS procedure as presented in~\citep{Ross2008} to the supplemental document.
RTBSS is a greedy algorithm that explores a lower-bound on the optimal $V(b)$ using a diversity of actions within a limited number of look-aheads.
It selects the policy that maximises this lower-bound.

We present the specialisation of the RTBSS planner to our setting.
At each decision point $t$, the algorithm calculates $Q(b_t, \stay)$, the Q-value of stopping and searching at the current node and an estimate for descending, $\hat{Q}(b_t, \go)$, obtained by assuming that the child nodes are leaf nodes, or equivalently, by taking the value of a second descent step, $Q(b_{t+1}, \go)$ to be 0.
In particular,
\begin{align*}
Q(b_t, \stay)
&= \textstyle b_{c_t}r(c_t) + (1 -  b_{c_t})(-1) = b_{c_t}(r(c_t) + 1) - 1 \\
\hat{Q}(b_t, \go)
&\triangleq \textstyle \sum_{c' \in \mathcal{C}(c_t)} p\left(o=c' \mid b_{t}, \go\right) Q(\tau(b_{t}, \go, o), \stay) \\
&= \textstyle \sum_{c' \in \mathcal{C}(c_t)} \guid\left(c_t, c'\right) (b_{c'}(r(c') + 1) - 1) \,.
\end{align*}

\begin{algorithm}
\caption{\textsc{Simplified RTBSS Policy specified for HQS}}
\label{alg:policy}
\begin{algorithmic}[1]
\STATE Initialise $b$ such that $b(\state{c_0}{1}) \gets 1$
\WHILE{\textsc{IsNotLeafBeliefState($b$)} and $\pi(b, \stay) = 0$}
  \STATE Compute $Q(b,\stay)$ and $\hat{Q}(b,\go)$
  \STATE $a^* \gets $ \textbf{if} $\hat{Q}(b,\go) > Q(b,\stay)$ \textbf{then} $\go$ \textbf{else} $\stay$
  \STATE $\pi(b, a^*) \gets 1$
  \STATE $b \gets \tau(b, a, o)$ \COMMENT{It focuses on the $o$ in the right path only}
\ENDWHILE
\STATE $\pi(b, \stay) \gets 1$ \COMMENT{The bot can only choose to search at a leaf node}
\end{algorithmic}
\end{algorithm}
Our simplified algorithm (in~\cref{alg:policy}) provides a belief-based policy for descending through the hierarchy, which at any given node determines whether to descend further or to stop.
The simplified algorithm helps HQS become less computationally demanding and achieves a polynomial time complexity.
\begin{proposition}
The worst case complexity of computing $\HQS$ for $N$ items is $O(N^3 \mathcal{F}(\similarity))$ where $\mathcal{F}(\similarity)$ is the complexity of the similarity function.
\end{proposition}
The sketch of the proof can be found in the supplemental document.

%% file: experiment.tex
\section{Experimental Study}
\label{sec:experiments}
We show how HQS works through a case study using a small dataset and compare it with an existing approach.
Then, we will discuss on scaling the HQS.

\subsection{Case Study on \texttt{Amazon}}
This study uses the textual information for the items from the \texttt{Amazon}\footnote{\url{http://jmcauley.ucsd.edu/data/amazon}}~\citep{mcauley2015image}, and bases the guidance function on a TF-IDF similarity score.
We select commonly seen daily life products and manually construct a number of hierarchies, varying their quality.
To ensure that the hierarchies can be intuitively assessed by inspection, we restrict to just 12 items.
This approach can help us quickly assess HQS procedure by assessing if it orders the hierarchies as expected.

\paragraph{The Selected Items}
Some TF-IDF details about the items are displayed in Table S1 and S2 (in the supplemental document).
We refer to items by their indices as specified in that table.
More details are also revealed in the supplemental document.

Since the data is represented by TF-IDF vectors, which we write as $\bm{v}_x$ for item $x$, we define the similarity function by
\begin{align*}
\similarity(x, c)
= \begin{cases}
1 & c = \{x\} \\
\frac 1 {|c \backslash \{x\}|} \sum_{x' \in c \backslash \{x\}} \cos(\bm{v}_x, \bm{v}_{x'}) & \mbox{otherwise}
\end{cases}\,.
\end{align*}
This similarity is in spirit similar to the average link of Agglomerative Clustering (AC)~\citep{day1984efficient}. {It is important to exclude $x$ itself from computing the similarities over a non-singleton cluster, since the maximum cosine similarity between any $(x, y)$ pair is only around $0.11$ in the selected items, and hence $\cos(\bm{v}_x, \bm{v}_x)$ would dominate the similarity score.}
For the guidance function in our experiments, we set $\delta=0.01$ which helps us increase the weight of the best cluster and ensure that the cluster with largest similarities to the item, has a dominant guidance value.

\begin{figure}[!t]
\centering
\subfloat[Hier-A]{
\includegraphics[width=0.295\textwidth]{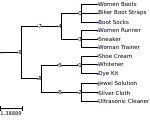}
\label{fg:hcA}}
~
\subfloat[Hier-B]{
\includegraphics[width=0.295\textwidth]{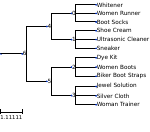}
\label{fg:hcB}} \\

\subfloat[Hier-C]{
\includegraphics[width=0.295\textwidth]{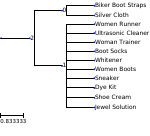}
\label{fg:hcC}}
~
\subfloat[Hier-D]{
\includegraphics[width=0.295\textwidth]{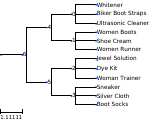}
\label{fg:hcD}}
~
\subfloat[Hier-E]{
\includegraphics[width=0.295\textwidth]{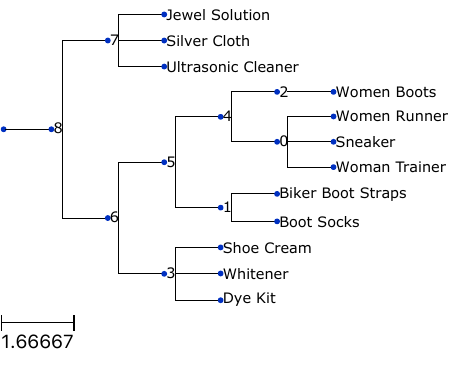}
\label{fg:hcE}}
\caption{Five different hierarchies}
\label{fg:hcExamples}
\end{figure}

\paragraph{Tested Hierarchies}
We analyse the HQS scores for five hierarchies in~\cref{fg:hcExamples}, where Hier-A is the ground-truth obtained from the labels provided in the \texttt{Amazon} dataset.
Hier-B and Hier-C are randomly generated hierarchies.
Hier-D is constructed to be deliberately poor.
All items in each leaf cluster have different ground-truth labels. For example, ``Whitener'', ``Biker Boot Straps'', and ``Ultrasonic Cleaner'' in cluster 0 of Hier-D are all in separate clusters according to the \texttt{Amazon} organisation.
Finally, Hier-E is constructed on purpose to show that a good hierarchy is achievable even different from the ground-truth.

\input{amazon_results_overall}

\subsubsection{Walk-through Analysis}
The second column of \cref{tb:amazon-overall} shows the HQS results of the five different hierarchies.
The ground truth hierarchy, A, provides the highest HQS score (0.5715), followed by a relatively high score for Hier-E as expected.
Hier-D provides the worst HQS score (-0.8477).
In this example, the HQS scoring scheme agrees with our intuition and we believe this ranking is reasonable.
\cref{tb:amazon-overall} also displays the values for each item obtained by RTBSS in the five hierarchies.
Next, we specify how the policy propagates step-by-step for those hierarchies.

\paragraph{Hier-A}
This hierarchy is generated given the ground truth labels of the items.
For instance, node 4 refers to ``Clothing, Shoes \& Jewelry'', node 7 refers to ``Women'', and 5 refers to ``Shoe Care \& Accessories'' etc., as shown in Table S1 and S2.

For item $x_0$, ``Women Boots'', staying and searching at the root earns an estimate $Q(b_0, \stay) = 0$.
Now, the similarities $x_0$ with the child nodes are $\similarity(x_0, c_7) = 0.061$ and $\similarity(x_0, c_8) = 0.0277$.
It follows that $\guid(c_9, c_7) = 0.9987$ and $\guid(c_9, c_8) = 0.0013$.
Clearly, both nodes have 6 items and so, the rewards for staying at both nodes are equal, $0.6226$.
It follows that
$\hat{Q}(b_0, \go)
= \sum_{o \in \{c_7, c_8\}} p(o \mid b, \go) Q(\tau(b_0,\go o))
= 0.6203$.
Since $Q(b_0,\stay)<\hat{Q}(b_0,\go)$, the bot performs action $\go$, and it randomly moves to the right node with probability $0.9986$.
From $c_7$ to $c_4$, it is a straight move as the estimates will be equal and the bot should explore a better chance for a higher value.
Let us review that cluster $c_0$ refers to ``Boots'' and $c_3$ refers to ``Fashion Sneakers'' under the category chain ``All $\rightarrow$ Clothing, Shoes \& Jewelry $\rightarrow$ Women'' ($c_9, c_7, c_4$).
It shows $\similarity(x_0, c_0)= 0.06, \similarity(x_0, c_3) = 0.061$ and we obtain $\guid(c_4,c_0) = 0.4158$ and $\guid(c_4,c_3)= 0.5829$.
This suggests that $x_0$ more strongly belongs to cluster $c_3$ than to $c_0$, the cluster it is assigned to.
Regarding $c_3$, its items ``Women Runner'', ``Women Sneaker'' and ``Women Trainer'' are highly close to ``Women Boots'' in some sense.
Upon computing the Q-values, the bot decides to stop at node $c_4$, since it is not confident that another descent step will lead to greater reward.
Finally, the oracle value $V_{x_0}$ is $0.6203$, a good score, but short of the maximum reward which captures the lack of clarity between the clusters at the lower levels of the hierarchy for this item.
This suggests that our measurement is rather reasonable in this case.

Another interesting example is $x_7$, ``Dye Kit''.
It starts navigating wrong at the first level.
The similarities $\similarity(x_7, c_7)$ and $\similarity(x_7, c_8)$ return $0.0367$ and $0.0297$ respectively.
It belongs to $c_8$ while the measurement prefers $c_7$.
In fact, ``Dye Kit'' is close to the shoe items in cluster $c_7$, but also close to the items
``Shoe Cleaner'' and ``Whitener'' in cluster $c_8$, as they are also cleaning tools (for jewellery).
As it stands, the bot will be guided to the wrong node and hence receives a negative reward.
It could well be the case that the situation would be improved with more data items.

The bot is not confident to descend to the bottom of the hierarchy for every case which is reflected in the associated score.
In summary, Hier-A is sufficiently good to help the bot descend and acquires a good HQS of $0.5715$.

\paragraph{Hier-B}
In this case,  the final HQS is $-0.2801$.
\cref{tb:amazon-overall} demonstrates that there are 6 items earning a negative reward as the hierarchy guides them to the ``wrong'' node.
Three more items earn a score of 0, as the bot gets confused at the root level and so it prefers searching at the root.
Interestingly, a random hierarchy does not necessarily achieve an absolute HQS of zero.
The HQS measurement is apparently not linear.

\paragraph{Hier-C}
Its HQS is $-0.3097$.
Similarly to Hier-B, there are 6 items earning a negative reward as the hierarchy has only two items achieving positive scores which are the two in the small cluster,
``Biker Boot Straps'' and ``Silver Cloth'' obtain $0.8892$ and $0.8789$ respectively.
The similarities are diluted compared with the small node, since the sibling clusters are too diversified.
However, HQS unveils the fact that the hierarchy is poor for item retrieval.
Although we compute the value for the bot searching for each item, HQS considers the hierarchy as a whole.
Thus, even a poor hierarchy might enable efficient search for a few items, while there are many more poorly scoring items.

\paragraph{Hier-D}
For this hierarchy, the overall HQS score of $-0.8477$,
reflects the fact that this is a hierarchy within which the bot has difficulty in searching.
HQS successfully reveals this by returning a score close to the minimum, the worst performing one among the exhibited hierarchies.
This implies that HQS correctly downgrades a random hierarchy.

\paragraph{Hier-E}
Finally, we construct a ``good'' hierarchy (\cref{fg:hcE}) while its style is very different to the ground-truth.
It separates the items about jewellery and the items associated with shoes at the first level.
Overall, there is only one item, ``Shoe Cream'' which receives a negative value $-0.7972$.
At the first decision point for this item, $\guid(c_8,c_7)$ is $0.8894$ which leads the bot to the wrong node with a dominant probability.
The item's textual representation tends to be better associated with the cleaning toolkit.
Interestingly, ``Biker Boot Straps'' receives the similarities $0.0379$ and $0.0375$ for $c_6$ and $c_7$ respectively; thus the bot stops at the root node.
It implies that the HQS acknowledges hierarchies with different structures to the ground-truth as long as the hierarchy is efficient for search.

\subsubsection{Comparison to Existing Approach}
We show that HQS outperforms HAI~\citep{Johnson2013} on our example hierarchy.

HAI defines a distance function $d(x, y, \tree) = |c_{x, y}| / {N}$ where $c_{x,y}$ is the closest ancestor node of items $x$ and $y$ when such an ancestor is not a leaf node; otherwise the distance returns $0$.
The HAI metric requires a ground-truth hierarchy and we denote it by $\tree_{gt}$.
Therefore, HAI is formulated as
\begin{align*}
\mathrm{HAI}(\tree, \tree_{gt}) = 1 - \frac 1 {N^2} \sum_{i=1}^N \sum_{j=1}^N \lvert d(x_i, x_j, \tree) - d(x_i, x_j, \tree_{gt}) \rvert \,.
\end{align*}
We conduct the evaluation over the hierarchies B--E with Hier-A as ground truth.
Their HAI scores are $0.64, 0.46, 0.58, 0.85$ respectively. Both HQS and HAI rank Hier-E and Hier-B as first and second, excluding the ground-truth. However, the HAI score of 0.58 for Hier-D, which was generated to serve as the worst example, is better than the score of Hier-C (which should not happen) and also numerically close to that of Hier-B.

\subsection{Discussion of Scaling HQS}
HQS can easily be run in parallel, since the POMDPs for the items are completely independent.
In this section, we demonstrate that sampling can further help with the runtime efficiency--with sufficient samples, one can achieve an approximated HQS close to the true HQS.
With the number of items sampled, the discrepancy between the sampled HQS and the true HQS decreases exponentially and the runtime speeds up linearly.

Again, we use \texttt{Amazon} but keep $37,826$ samples and $100$ dimensions with PCA.
The hierarchy is simply generated using AC and contains $75,651$ nodes with a maximum height $65$.
The prototype of HQS is implemented with Python.\footnote{The experiments are conducted in a computer equipped with AMD Ryzen 7 1700X@2.2GHz 8-Core Processor and 64GB of RAM, using 16 threads. See \url{https://github.com/weipeng/pyhqs} for the code.}
Tree is a relatively memory-intense data structure and the data size is also a consideration.
Hence, RAM prevents us from examining larger (in another order of magnitude) trees.

\begin{figure}[!ht]
\begin{minipage}{.38\textwidth}
\centering
\includegraphics[scale=.46]{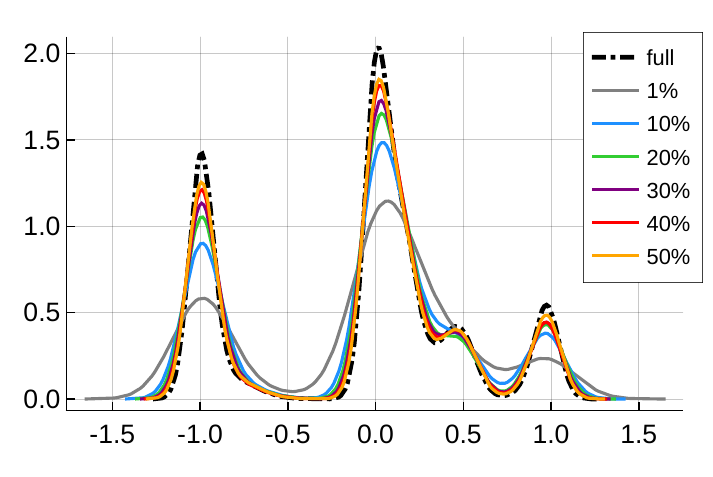}
\caption{Empirical density}
\label{fg:hqs-cdf}
\end{minipage}
~
\begin{minipage}{.28\textwidth}
\centering
\includegraphics[scale=.48]{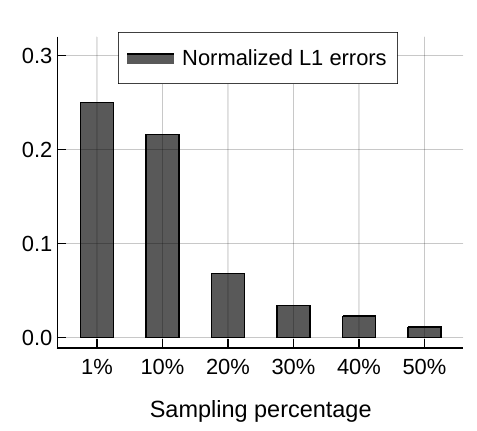}
\caption{Normalised $L_1$ errors}
\label{fg:sampled_hqs}
\end{minipage}
~
\begin{minipage}{.28\textwidth}
\centering
\includegraphics[scale=.48]{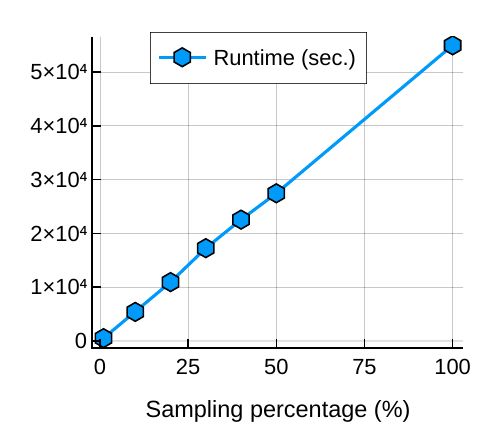}
\caption{Runtime of the sampled cases}
\label{fg:runtime}
\end{minipage}
\end{figure}

Let us now down-sample the items for computing the oracle values.
\cref{fg:hqs-cdf} displays the empirical densities of the HQS for the whole set of \texttt{Amazon} items, fitted using \texttt{StatsPlots}\footnote{\url{https://github.com/JuliaPlots/StatsPlots.jl}}.
The black dashed line is the true HQS with all items computed, while the others have sampled the corresponding percentages of items.
It is apparent that the density curve with more samples gets closer to the one of the entire set of data.
\cref{fg:sampled_hqs} exhibits the $L_1$ errors between the true HQS and the sampled HQS, which are divided by the absolute true HQS.
It is clear that the errors decrease drastically as more items are sampled.
After $30\%$ sampling, the errors are less than $5\%$ of the absolute value of the true HQS.
As expected, the HQS of the cases with fewer samples are less accurate.
Finally, the runtime is plotted in~\cref{fg:runtime}, and is shown to follow approximately a linear relation.

Overall, HQS can benefit from that it is completely parallelisable.
In addition, sampling techniques can also help improve the efficiency.
We would like to emphasise that the runtime efficiency is apparently improvable with better programming code and machines.

%% file: amazon_results_overall.tex
\begin{table}[!t]
\footnotesize
\centering
\caption{HQS results of the five hierarchies (A--E in~\cref{fg:hcExamples}) and the value $V$ for the twelve items (indexed from 0--11) in each hierarchy.}
\label{tb:amazon-overall}
\begin{tabular}{c a c c c c c c c c c c c c}
 &   HQS & 0 &   1 &   2 &   3 &   4 &   5 &   6 &   7 &   8 &   9 &   10 &   11 \\
\midrule
A & {.572} & .620 & .747 & {.827} & {.811} & .600 & {.518} & .793 & -.679 & {.562} & {.432} & {.812} & {.817} \\
B & -.280 &
.0 &
-.991 &
-.989 &
{.863} &
-.955 &
-.736 &
.760 &
.659 &
-.995 &
.0 &
.749 &
.0 \\
C & -.310 &
-.752 &
-.999 &
.0 &
.0 &
.0 &
 {.889} &
 {.857} &
-.933 &
 {.879} &
.0 &
-.965 &
-.979 \\
D &  {-.848} &
-.668 &
-.740 &
-.708 &
-.996 &
-.652 &
 {-1.00} &
-.791 &
-.607 &
-.995 &
-.925 &
-.990 &
-.901 \\
E & .327 &
.611 &
-.797 &
.673 &
.0 &
 {.533} &
.0 &
 {.834} &
.0 &
.423 &
.0 &
 {.835}&
 {.808}\\
\bottomrule
\end{tabular}
\end{table}


%% file: conclusion.tex
\section{Conclusion}
\label{sec:conclusion}

We have proposed HQS, an approach for assessing the quality of hierarchical clusters which needs no ground-truth information.
It employs POMDP for modelling the uncertainty and the decision making for searchers with regard to search and navigation in a hierarchy, and extends the model to measure the quality of the hierarchies.
The experiments show that HQS can order hierarchies reasonably based on their qualities without the ground-truth, and perform better than a state-of-the-art approach which requires the ground-truth hierarchy.
For future work, we would like to continue exploring better models and policies, and analyse the properties of the measure theoretically.

%% file: appendix.tex
\begin{appendices}

\section{Example of A POMDP for a Hierarchy}
Consider a search over the three-node hierarchy with only three nodes, the root node $c_0$ and its two children $c_1$ and $c_2$. It contains eight possible states:
\[
\state{{c_0}}{1}, \state{{c_0}}{0}, \state{{c_1}}{1}, \state{{c_1}}{0}, \state{{c_2}}{1}, \state{{c_2}}{0}, \state{\emptyset}{1}, \state{\emptyset}{0} \,.
\]
The POMDP for this simple tree yields belief states $b_{c_0}$, $b_{c_1}$, $b_{c_2}$ when the bot is at the corresponding node, and the trivial belief states at the fully observed terminal states $\state{\emptyset}{1}$ and $\state{\emptyset}{0}$.
The bot moves using the guidance function values $\tilde{\eta} \triangleq \eta({c_0},{c_1})$ and $\eta({c_0},{c_2}) = 1 - \tilde{\eta}$ to determine the next node when the action $\go$ is selected.
The set of reachable belief states is represented in an AND-OR tree in~\cref{fg:simple_andor_tree} (see a similar figure in~\citep{Ross2008}).
In this figure, an action must be chosen at an OR node, the choice of which leads to the set belief states, over all possible observations, that must be considered at the AND nodes.
Expected rewards, $R(b,a)$ are represented on the arcs from OR- to AND-nodes, while the transition probabilities $p(o \mid b,a)$ are represented on the arcs from AND- to OR-nodes.
Working from the leaf nodes back to the root, we can read from the tree that action $\stay$ at node ${c_0}$, would lead to an expected reward of
\[
  Q(b_{c_0},\stay) = b_{c_0} r({c_0}) + (1 - b_{c_0})(-1) = b_{c_0}( r({c_0}) + 1) - 1
\]
while action $\go$ at ${c_0}$ would lead to an expected reward of
\begin{align*}
Q(b_{c_0},\go)
&=b_{c_0}\left[\tilde{\eta}^2(r({c_1})+1) + (1 - \tilde{\eta})^2(r({c_2})+1)\right] - 1
\end{align*}
where the expression comes from $b_{c_1} = \tilde{\eta}b_{c_0}$ and $b_{c_2} = (1-\tilde{\eta})b_{c_0}$ via~\cref{eq:beliefupdate}.
\begin{align}
\label{eq:beliefupdate}
  b'\left(\state{c'}{1}\right)
  &= \eta(c, c') b\left(\state{c}{1}\right) \qquad  b'\left(\state{c'}{0}\right) = 1 - \eta(c, c') b\left(\state{c}{1}\right) \, .
\end{align}

\input{example_or_tree}

\section{Insights into the Guidance Function}
Recall that the guidance function is
\begin{equation}
  \label{eq:eta}
  \eta(c, c',x) = p(x \in c' \mid x \in c) \triangleq
  \frac {\exp\{\similarity(x, c')/\delta\}} {\sum_{c'' \in \mathcal{C}(c)} \exp\{\similarity(x, c'')/\delta\}} \, .
\end{equation}
This can be any kind of similarity that is used in measuring clustering results.
Leaving the choice of similarity function open adds flexibility to the measure by allowing the users to customise the comparison for various hierarchies of specific datasets.
A simple choice is the inverse or the negative of Euclidean distance, assuming items can be mapped to points in $\mathbb{R}^n$; Bayesian models could use a similarity based on a distribution density; if items are represented as Term Frequency and Inverse Document Frequency (TF-IDF) vectors of textual data, then cosine similarity might be appropriate; and so on.
The $\eta$ function should return a higher probability score for the node $c'$ that is ``closest'' to $x$ among the siblings.
The parameter $\delta$ is a \emph{temperature} parameter in the Boltzmann function.

Suppose there are two child clusters $A$ and $B$ of parent $C$ such that $\similarity(x, A) < s_\epsilon$ and $\similarity(x,B) < s_\epsilon$  for some $s_\epsilon \approx 0$, and yet $\similarity(x,A) \ll \similarity(x,B)$.  For example, suppose the similarities of $x$ to random cluster $A$ and cluster $B$, respectively, return $1e-50$ and $1e-70$.
It would be better to have $\eta(C,A) \approx \eta(C,B)$, rather than $\eta(C,A) \approx 1$, for two such clusters.
Deep in the hierarchy, the bot should only choose to descend to a cluster when the evidence that it contains the target is strong.
A $\delta$ parameter that increases with the depth of the tree ensures that the similarity values between two clusters must be increasingly more distinct deeper in the tree before one cluster is preferred over another.

Thus, $\delta_t$ can be defined as a function over the depth, $t$, of the hierarchy, s.t. $\delta_t \triangleq \delta \nu^{t}$ where $\nu \in [1, \infty)$.
Setting $\nu=1$ makes $\delta_t$ invariant with depth.

\section{Policy}
This section is devoted to two parts.
The first is to show the complete version of the Real-Time Belief Space Search (RTBSS).
Then, we analyse the complexity.

\subsection{RTBSS}
Algorithm~\ref{alg:rtbss} demonstrates the original RTBSS procedure as presented in~\citep{Ross2008}.
It heavily relies on the function $\textsc{Expand}(b, a)$ in Algorithm~\ref{alg:expand} to explore the POMDP.
The Boolean function $\textsc{IsLeaf}(b)$ returns \texttt{true} if the only belief states reachable from $b$ with non-zero probability are the terminal states.
RTBSS is a greedy algorithm that explores a lower-bound on the optimal $V(b)$ using a diversity of actions within a limited number of look-aheads and selects the policy that maximises this lower-bound.
Considering that the look-aheads are capped, this can also be described as a myopic policy and so follows
the proposal in~\citep{fern2007decision,Eugene2019} to use myopic heuristics for approximating the  $Q$ value for each belief-action pair to alleviate the intractable computations in a POMDP.
\begin{algorithm}[!ht]
\caption{\textsc{RTBSS}} \label{alg:rtbss}
\begin{algorithmic}[1]
\REQUIRE $d$, the maximum look-aheads which is fixed to $2$ in our settings
\ENSURE  $\pi$, the policy function
\STATE Initialise $b$
\REPEAT
  \STATE $L, a \gets \textsc{Expand}(b, d)$
  \STATE $\pi(b, a) \gets 1$
  \STATE Execute $a$ and perceive $o$
  \STATE $b \gets \tau(b, a, o)$
\UNTIL{\textsc{IsLeaf($b$)}}
\end{algorithmic}
\end{algorithm}

\begin{algorithm}[!ht]
\caption{\textsc{Expand}} \label{alg:expand}
\begin{algorithmic}[1]
\REQUIRE $b$, the current belief state
\REQUIRE $d$, the number of levels to explore, must be $\ge 0$
\ENSURE $L^*$, optimal lower bound
\ENSURE $a^*$, optimal action
\STATE $L^* \gets -\infty$
\IF{$d = 0$ or \textsc{IsLeaf}($b$)}
  \STATE$L(a) \gets R(b, a)$
\ELSE
  \FOR{$a \in A$}
    \STATE $L(a) \gets R(b, a) + \gamma \sum_{o \in O}p(o \mid b, a)\textsc{Expand}(\tau(b, a, o), d-1)$
  \ENDFOR
\ENDIF
\STATE $L^* \gets \max_a L(a)$
\STATE $a^* \gets \argmax_a L(a)$
\end{algorithmic}
\end{algorithm}


%

\subsection{Time Complexity of HQS}
\begin{proposition}
The worst case complexity of computing HQS for $N$ items is $O(N^3 \mathcal{F}(\similarity))$ where $\mathcal{F}(\similarity)$ is the complexity of the similarity function.
\end{proposition}

\begin{proof}[Sketch]
Solving a finitely horizontal POMDP is PSPACE-complete~\citep{papadimitriou1987complexity}, while luckily it does not apply to our case. Consider the HQS calculation for a single target $x$. Let $\mathcal{F}(\similarity)$ represent the complexity of calculating the similarity between $x$ and a cluster.
Let us reasonably assume that each non-root node has at least one sibling in the hierarchy.
For the data with $N$ entries and corresponding hierarchy with $M^*$ nodes, the maximum $M^*$ is $2N-1$.
This holds when the tree splits one data point as a leaf and all others remain as one cluster, until all points become leaves.

Least optimally, the searcher needs to estimate the return at all nodes for a certain target, which will be in $O(M^*)$.
However, the policy can still be pruned as reaching a wrong node  finally receives the reward $-1$.
Accordingly the reward computation can concentrate on the path wherein each node contains the target.
Denote the number of children of the $t^{th}$ parent in the right path by $N_t$.
The complexity will then follow $O(\sum_{t} {N}_{t}) = O(M^*N)$ which is thus $O(N^2)$.
Hereafter, the {guidance function} requires $O(N^2 \mathcal{F}(\similarity))$ computations given that $O(\mathcal{F}(\similarity))$ is the complexity for calculating the similarity for a data point to a cluster---which will be polynomial for most commonly used similarity choices.
\end{proof}

Nevertheless, the average case for the height of a tree is always logarithmic.
We can write the average complexity of searching for a target as $O(a \log_a M) = O(a \log_a N)$ where $a$ is a constant for the number of children.
The average case for the HQS is therefore $O(a \log_a N \cdot N \mathcal{F}(\similarity)) = O(N \log N\mathcal{F}(\similarity))$.
The polynomial result concludes that HQS is practically applicable.

\section{Experimental Setup}
\subsection{For the Implementation of HQS}
We show some numerical details about the items in Table~\ref{tb:amazon1} and \ref{tb:amazon2}.
We refer to items by their indices as indicated in the tables.
The left column is the title of the item, and the second column contains the top 10 terms in the title and the description of the item, with the corresponding TF-IDF score in parentheses.
As the examples all come from the large fashion category, the TF-IDF score is computed only on this subset of the \texttt{Amazon} data.
However, when computing the similarities, given the small number of items, we still consider all the features without any feature selection techniques .

\subsection{For Scaling the Approach}
For the two experiments, we adopt the following similarity for use in the guidance function
\begin{align*}
\mathcal{S}({x}, c) = \frac 1 {\| \bm{v}_x - \bar{\bm{v}}_{c}\|^2 + 0.0001} \, .
\end{align*}

For \texttt{Amazon} after PCA applied, we set $\delta = \ceil{\frac 1 {100}} = 0.01$ where $100$ is the number of dimensions that we keep.

\begin{table}[ht]
\renewcommand{\thetable}{S\arabic{table}}
\input{amazon_table_1}
\caption{Amazon item scores 1}
\label{tb:amazon1}
\end{table}

\begin{table}[ht]
\renewcommand{\thetable}{S\arabic{table}}
\input{amazon_table_2}
\caption{Amazon item scores 2}
\label{tb:amazon2}
\end{table}

\end{appendices}

%% file: example_or_tree.tex
\begin{figure}[t]
\footnotesize
\centering
\begin{tikzpicture}[scale=0.9, every node/.style={scale=0.9}]

\draw (4,3.75) -- (5,3.75) -- (4.5, 3.75+.9) -- cycle;
\node[] at (4.5 ,3.75+.3) {$b_{c_0}$};
\node[] at (2.5 ,3.65) {$b_{c_0}(1+r({c_0}))-1$};
\node[] at (5.8 ,3.65) {0};

\draw[] (1, 2.75) circle (.5) node[] {$a_s$}; 
\draw[] (8, 2.75) circle (.5) node[] {$a_d$}; 
\node[] at (0, 2) {$1-b_{c_0}$};
\node[] at (2.2-.5, 2) {$b_{c_0}$};
\node[] at (6.3, 2) {$\eta({c_0}, {c_1})$};
\node[] at (9.8, 2) {$\eta({c_0}, {c_2})$};

\draw (-.5,0.5) -- (.5,0.5) -- (0, .9+.5) -- cycle; 
\draw (1.5,0.5) -- (2.5,0.5) -- (2, .9+.5) -- cycle; 
\node[] at (0.03, 0.8) {$b_{\phi, 0}$};
\node[] at (2.03, 0.8) {$b_{\phi, 1}$};

\draw (5,0.5) -- (6,0.5) -- (5.5, 1.4) -- cycle; 
\draw (10,0.5) -- (11,0.5) -- (10.5, 1.4) -- cycle; 
\node[] at (5, 1+.5) {$o = {c_1}$};
\node[] at (11, 1+.5) {$o = {c_2}$};
\node[] at (5.53, .8) {$b_{c_1}$};
\node[] at (10.53, .8) {$b_{c_2}$};

\node[] at (4., -.1) {$b_{c_1}(1+r({c_1}))-1$};
\node[] at (9., -.1) {$b_{c_2}(1+r({c_2}))-1$};

\draw[] (5.5,-.75) circle (.5) node[] {$a_s$}; 
\draw[] (10.5,-.75) circle (.5) node[] {$a_s$}; 
\node[] at (4.4, -1.25) {$1-b_{c_1}$};
\node[] at (6.3, -1.25) {$b_{c_1}$};
\node[] at (9.4, -1.25) {$1-b_{c_2}$};
\node[] at (11.3, -1.25) {$b_{c_2}$};

\draw (3.5,-2.5) -- (4.5,-2.5) -- (4., .9-2.5) -- cycle; 
\draw (6.5,-2.5) -- (7.5,-2.5) -- (7, .9-2.5) -- cycle;
\draw (8.5,-2.5) -- (9.5,-2.5) -- (9, .9-2.5) -- cycle;
\draw (11.5,-2.5) -- (12.5,-2.5) -- (12, .9-2.5) -- cycle;
\node[] at (4.53-.5, 0+.3-2.5) {$b_{\phi, 0}$};
\node[] at (4.53+2.5, 0+.3-2.5) {$b_{\phi, 1}$};
\node[] at (9.53-.5, 0+.3-2.5) {$b_{\phi, 0}$};
\node[] at (9.53+2.5, 0+.3-2.5) {$b_{\phi, 1}$};

\draw[>=latex, ->, postaction={draw}] (4.5, 3.75) -- (1.5, 2.75);
\draw[>=latex, ->, postaction={draw}] (4.5, 3.75) -- (7.5, 2.75);
\draw[>=latex, ->, postaction={draw}] (1, 2.25) -- (0, .9+.5);
\draw[>=latex, ->, postaction={draw}] (1, 2.25) -- (2, .9+.5);
\draw[>=latex, ->, postaction={draw}] (8, 2.25) -- (5.5, .9+.5);
\draw[>=latex, ->, postaction={draw}] (8, 2.25) -- (10.5, .9+.5);
\draw[>=latex, ->, postaction={draw}] (5.5, .5) -- (5.5,-.25);
\draw[>=latex, ->, postaction={draw}] (10.5, .5) -- (10.5,-.25);

\draw[>=latex, ->, postaction={draw}] (5.5,0-1.25) -- (4, .9-2.5);
\draw[>=latex, ->, postaction={draw}] (5.5,0-1.25) -- (7, .9-2.5);
\draw[>=latex, ->, postaction={draw}] (10.5,0-1.25) -- (9, .9-2.5);
\draw[>=latex, ->, postaction={draw}] (10.5,0-1.25) -- (12, .9-2.5);
\end{tikzpicture}
\caption{An AND-OR tree of reachable belief states from node ${c_0}$ of the three-node hierarchy}
\label{fg:simple_andor_tree}
\end{figure}
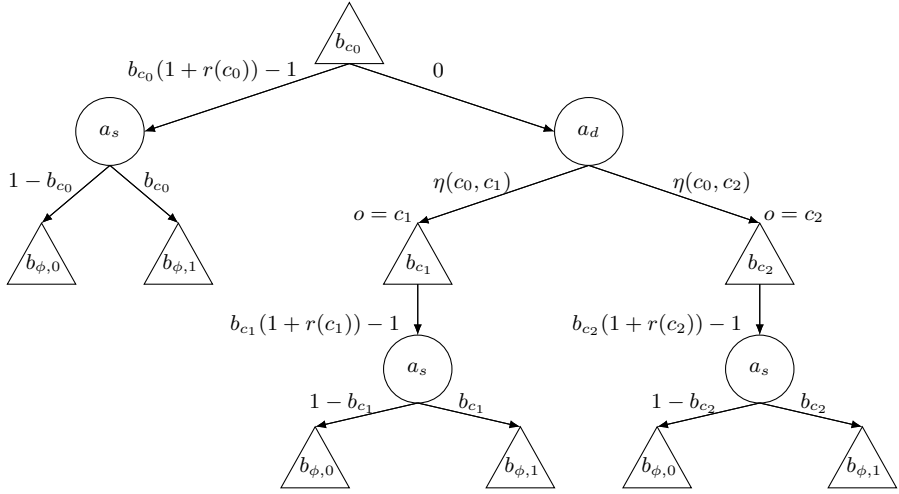

%% file: amazon_table_1.tex
\begin{tabular}{l  p{3cm} p{10cm}}
\toprule
 index &          short name &                                                                                                                                                                                  top 10 terms \\
\midrule
 0 &  Women Boots &  Harness (0.2455), term (0.2186), long (0.1984), Womens (0.181), durability (0.1709), boots (0.1647), inch (0.1547), Crushed (0.1514), element (0.1465), tougher (0.1465),  \\
 1 &  Shoe Cream &  Meltonian (0.2913), waxes (0.2768), cream (0.2169), cloth (0.1914), afterwards (0.1653), Misc (0.158), staining (0.1489), honored (0.1489), creamy (0.1489), terrific (0.1489),  \\
 2 &  Women Runner &  Ascend (0.4741), Wave (0.3866), MIZUNO (0.237), EU (0.2266), SZ (0.2135), lends (0.2089), Mizuno (0.2049), UK (0.1822), Running (0.1822), China (0.1659),  \\
 3 &  Whitener &  Whitener (0.5887), Sport (0.3376), chalky (0.2943), restores (0.2502), KIWI (0.2324), formula (0.218), scuffs (0.218), polish (0.1974), covers (0.1974), Kiwi (0.1925),  \\
 4 &  Sneaker &  Coach (0.4753), signature (0.2487), leather (0.2173), preeminent (0.1759), Poppy (0.1759), Barrett (0.1759), emerged (0.1759), resulting (0.1682), Scribble (0.1682), coach (0.1682),  \\
 5 &  Biker Boot Straps  &  to (0.2226), 6in (0.2192), are (0.2111), clips (0.1902), in (0.1884), Straps (0.188), sold (0.1716), prevent (0.1564), Boot (0.15), SP6 (0.1402),  \\
\bottomrule
\end{tabular}

%% file: amazon_table_2.tex
\begin{tabular}{l  p{3cm} p{10cm}}
\toprule
 index &           short name &                                                                                                                                                                                               top 10 terms \\
\midrule
 6 &  Jewel Solution &  cleaning (0.2535), components (0.2389), metals (0.2297), precious (0.1925), solution (0.1867), free (0.1512), accumulate (0.1482), Biodegradable (0.1482), titanium1 (0.1482), injectors (0.1482),  \\
 7 &  Dye Kit &  TRG (0.4005), included (0.2872), Turquoise (0.2277), Detailed (0.2141), Everything (0.2117), coats (0.2105), dye (0.1958), instructions (0.1937), Self (0.1912), Dye (0.1852),  \\
 8 &  Silver Cloth &  silver (0.3919), tarnishing (0.3315), tarnish (0.189), Silvershield (0.1713), Cadet (0.1713), shining (0.1414), drawer (0.1414), trade (0.1389), Tarnish (0.1389), will (0.1277),  \\
 9 &  Woman Trainer  &  Ryka (0.5414), Rythmic (0.406), Womens (0.1547), the (0.148), Athena (0.1353), cardio (0.1353), fittest (0.1353), Rhythmic (0.1353), kickboxing (0.1353), gain (0.1294),  \\
 10 &  Ultrasonic Cleaner  &  Professional (0.3063), ultrasonic (0.2914), grade (0.2771), NUMWPT (0.2082), Qt (0.2082), Joy4Less (0.2082), automotive (0.1925), transducer (0.1875), Heater (0.1875), controls (0.1834),  \\
 11 &  Boot Socks  &  dead (0.3354), Cuffs (0.3036), gorgeous (0.2605), drop (0.2396), lace (0.2396), season (0.2363), Add (0.2348), cuffs (0.234), put (0.2248), Socks (0.2194),  \\
\bottomrule
\end{tabular}